\documentclass{article}
\usepackage{spconf,amsmath,graphicx, booktabs, caption, subcaption, hyperref}

\usepackage[table,x11names]{xcolor}
\usepackage{multirow}

\newcommand{\specialcell}[2][c]{%
  \begin{tabular}[#1]{@{}c@{}}#2\end{tabular}}
\definecolor{shadecolor}{rgb}{0.94,0.94,0.94}
\definecolor{noshadecolor}{rgb}{1.0,1.0,1.0}

\title{SCALABLE MULTI-DOMAIN DIALOGUE STATE TRACKING}
\name{Abhinav Rastogi, Dilek Hakkani-T{\"u}r, Larry Heck}
\address{Google Research, Mountain View}
\begin{document}
%
\maketitle
\begin{abstract}
Dialogue state tracking (DST) is a key component of task-oriented
dialogue systems. DST estimates the user's goal at each user
turn given the interaction until then.
State of the art approaches for state tracking rely on deep
learning methods, and represent dialogue state as a distribution over
all possible slot values for each slot present in the ontology.  Such
a representation is not scalable when the set of possible values are
unbounded (e.g., date, time or location) or dynamic (e.g., movies or
usernames).
Furthermore, training of such models requires labeled data, where each
user turn is annotated with the dialogue state, which makes building
models for new domains challenging.  In this paper, we present a
scalable multi-domain deep learning based approach for DST. We introduce a novel
framework for state tracking which is independent of the slot value set, and
represent the dialogue state as a distribution over a set of values of interest
(candidate set) derived from the dialogue history or knowledge. Restricting
these candidate sets to be bounded in size addresses the problem of
slot-scalability.
Furthermore, by leveraging the slot-independent architecture and transfer
learning, we show that our proposed approach facilitates quick
adaptation to new domains.
\end{abstract}
\section{Introduction}
\label{sec:intro}

Dialogue state tracking (DST) is a core task in task-oriented dialogue systems.
The dialogue state (DS) represents the system's estimate of the user's goal
given the conversation history and is used (1) to make calls to an API, database
or action provider based on the user's preferences specified during the dialogue
(2) as an input to the dialogue
policy~\cite{henderson2013deep,shah2016interactive} which predicts the next
system response. Figure \ref{fig:dialogue-example} shows an example dialogue
with dialogue state annotations after each turn. Having a separate component for
DST prevents the dialogue system from having to capture long term dependencies
between raw natural language utterances. The Dialogue State Tracking Challenges
(DSTC) provide a common experiment framework with annotated human-machine
dialogue data sets~\cite{DSTC1, DSTC2, DSTC3, DSTC4, DSTC5}.

The state of the art models for DST use deep learning methods to predict the DS
based on the conversation history~\cite{henderson2013deep, henderson2014,
  mrksic2015, mrksic2016, wen2016, hori2016dialog, dernoncourt2017robust,
  BingLiu2017}. These approaches use deep neural networks to learn
representations for user and system utterances as well as instantiations of
slot names and values in them.
In such approaches, an ontology defines the set of slots for a task
and the set of values associated with these slots. The dialogue state is
represented as a distribution over the value set for each slot, pre-specified
in the ontology. Some of the approaches also use the ontology entries to detect
possible slot values in the user utterances~\cite{henderson2014, wen2016}. In
practice, it is difficult or impractical to limit the ontology.
The number of possible values for a slot could be large or unbounded, making
scalability of these approaches a significant issue. Furthermore, such a
representation cannot deal with entities which are not seen during training,
making it difficult to work with dynamically changing databases.


\begin{figure}[t]
  \begin{tabular}{ l l}
\textbf{User:} & Book me a table for two at Cascal.\\
 &\texttt{restaurant=Cascal,\#people=two}\\
\textbf{System:} & I found a table for two at Cascal at 6 pm.\\
  & Does that work?\\
\textbf{User:} & 6 pm isn't good for us. How about 7 pm?\\
&\texttt{restaurant=Cascal,\#people=two,}\\
&\texttt{time=7 pm}\\
  \end{tabular}
  \caption{A dialogue with dialogue states after each turn.}
  \label{fig:dialogue-example}
\end{figure}

This paper describes a deep learning based approach to dialogue state
tracking that can represent slots with large or unbounded sets of possible
values. To tag slots in user utterances, we use multi-domain
language understanding (LU) models based on bi-directional recurrent
neural networks~\cite{dilek:2016, jaech:2016,
  ankur:2017}. Multi-domain
training~\cite{williams2013multi,mrksic2015} benefits from sharing
labeled data as well as slots across domains. The language
understanding module outputs are used to \emph{delexicalize} the user
utterances, which are processed by the DST for feature
extraction.  We then integrate a separate candidate generation step that
estimates a set of slot value candidates using the local conversation
context, as well as possibly external knowledge sources. DST operates only on
these candidates, resulting in an
approach scalable to large and rich datasets.  We introduce a novel
framework for state tracking, which extracts a rich set of features
that are independent of the slot value set.  To capture long term
dependencies in natural language, we investigate representing input
utterances using bi-directional recurrent neural networks with GRU
cells, extending previous work using deep feed-forward or
convolutional neural networks~\cite{henderson2014,shi2016}. We extract
slot value related features from the GRU cell output at positions
corresponding to the tags from the LU. Furthermore, for each slot and
candidate value pair, we extract features that check presence in the
history, independent of the slot and the value.

The final contribution of our work is an approach to share
parameters across different slots in a given domain and transfer
the parameters to a previously unseen dataset/domain. This removes
the requirement to train a model for each slot type in each domain and
facilitates quick addition of new slots to a domain. Furthermore, due to
the slot type or value independence of the used features, the proposed
approach simplifies integration of new domains to the dialogue
system.

In the next section, we describe candidate set generation for dialogue
state tracking that enables scaling to large slot value sets.  Section
3 details our approach, describing the set of input features that
enable scaling to new slot types and parameter sharing which enables
transfer learning to new domains. Finally, in Section 4, we describe
experimental results with DSTC2 dataset as well as new movie and
restaurant domain datasets we formed for slot scalability and transfer learning
experiments.

\section{The Dialogue State}
The dialogue state (or belief state) is a full representation of the
system's belief of the user's goal. Discriminative approaches
~\cite{williams2014web, henderson2014, mrksic2015, mrksic2016,
  wen2016} have been popular for DST. Many of these approaches model
the dialogue state as a joint distribution across all slots and make a
simplifying assumption of factoring the joint distribution as a
product of a distribution for each slot~\cite{henderson2014,
  mrksic2015, mrksic2016, wen2016}. Since the slot distributions are
over the set of all values taken by a slot, they further assume that
the complete list of all possible values of a slot is known. Such a
representation is not scalable when the dialogue system is to be
deployed for tasks with large databases which have a large number of
possible values for a slot or when the underlying database is dynamic
(e.g., movie, song, book databases). Furthermore, for common slots
like date and time, it is difficult to detail a complete list of
values of reasonable size. To handle these scalability issues, we
utilize the concept of candidate set for a slot.

\subsection{Candidate Set} \label{candset} The candidate set for a slot is
defined to be a set of values of that slot, along with associated scores. These
scores represent the degree of user's preference for the corresponding value in
the current dialogue. Let $C^t_s$ be the candidate set at the dialogue
turn\footnote{Defined to be a system utterance and the following user
utterance.} at index $t$ for a slot $s$ in a domain $D$ handled by the dialogue
system. At the start of the conversation, $C_s^0$ is empty for every slot. We
impose the bound $|C_s^t| \leq K$ to restrict the maximum possible size of a
candidate set. The initialization steps for the candidate set are listed below.
These steps ensure that a newly mentioned value is always included in the
candidate set and less preferred values are removed if needed.
\begin{enumerate}
\item Add all values associated with slot $s$ in the user utterance at turn $t$
      to $C_s^t$. These values are obtained from LU output.
\item Add all values associated with slot $s$ in the system utterance at turn
      $t$ to $C_s^t$. These values are obtained from the system actions.
\item Sort the candidates from candidate set in the previous turn,
      $c_{s, i}^{t-1} \in C_s^{t-1}$ in decreasing order of score and extend
      $C_s^t$ with $c_{s, i}^{t-1}$ until $|C_s^t| \leq K$.
\end{enumerate}

After the initialization step, DST predicts the
scores for each of candidate by using the score of the
candidate at the previous turn and the features from the conversation context.
Figure \ref{fig:fig1} shows the candidate sets for two slots considered at
the last turn of the example dialogue in Figure
\ref{fig:dialogue-example}, \{\emph{6 pm, 7pm}\} and \{\emph{Cascal}\}
for slots \emph{time} and \emph{restaurant}, respectively.

We want to highlight two points: (i) The initialization steps
for candidate sets shown above can be easily
extended to include candidates from external sources like slot values from
multiple ASR hypotheses or backend/API call responses. (ii)  The maximum
capacity of the candidate sets ($K$) should be large enough to ensure recently
introduced or highly probable values are not flushed out. In our experiments, we
have found that candidate sets rarely reach the maximum capacity with $K = 7$.

The idea of a candidate set is in spirit similar to summary state
introduced by~\cite{williams2005scaling} for dealing with
intractability issues resulting from large set of possible values for
a slot in partially observable Markov decision processes (POMDP)
approach for dialogue management. Summary POMDP keeps track of two
values for each slot, the best and the rest, whereas candidate sets maintain a
distribution over a set $K$ of values that could be determined from
the dialogue context as well as external resources. Our work
investigates this idea for scaling neural network based dialogue state trackers.


\subsection{State Representation}
We retain the slot independence assumption and factor the dialogue
state as a product of distributions for each slot. However, instead of
defining these distributions over $V_s$ (set of all possible values
for slot $s$), we restrict the distribution at turn $t$ to be over
${V'}_s^t = C^t_s \cup \{\delta_s, \phi_s\}$, where $C^t_s$ represents
the candidate set generated for slot $s$ at turn $t$, $\delta_s$
represents the \emph{dontcare} value (i.e., user has no preference for
slot $s$) and $\phi_s$ represents the \emph{null} value (i.e., slot
$s$ is not specified yet). This is a good approximation because the
values which have never been mentioned in the dialogue will have their
probability close to zero and hence don't need to be explicitly stored
in the dialogue state.

\begin{figure}[t]
  \centering
  \begin{subfigure}[b]{0.5\textwidth}
        \includegraphics[width=0.93\textwidth]{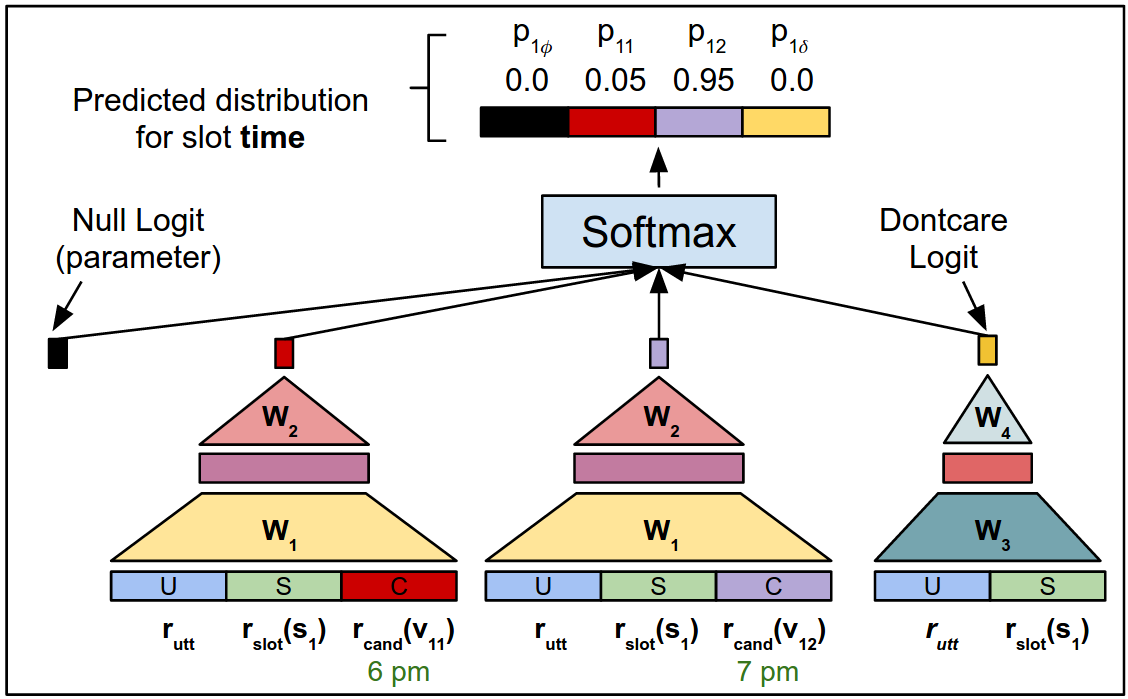}
  \end{subfigure}
  \begin{subfigure}[b]{0.5\textwidth}
        \includegraphics[width=0.93\textwidth]{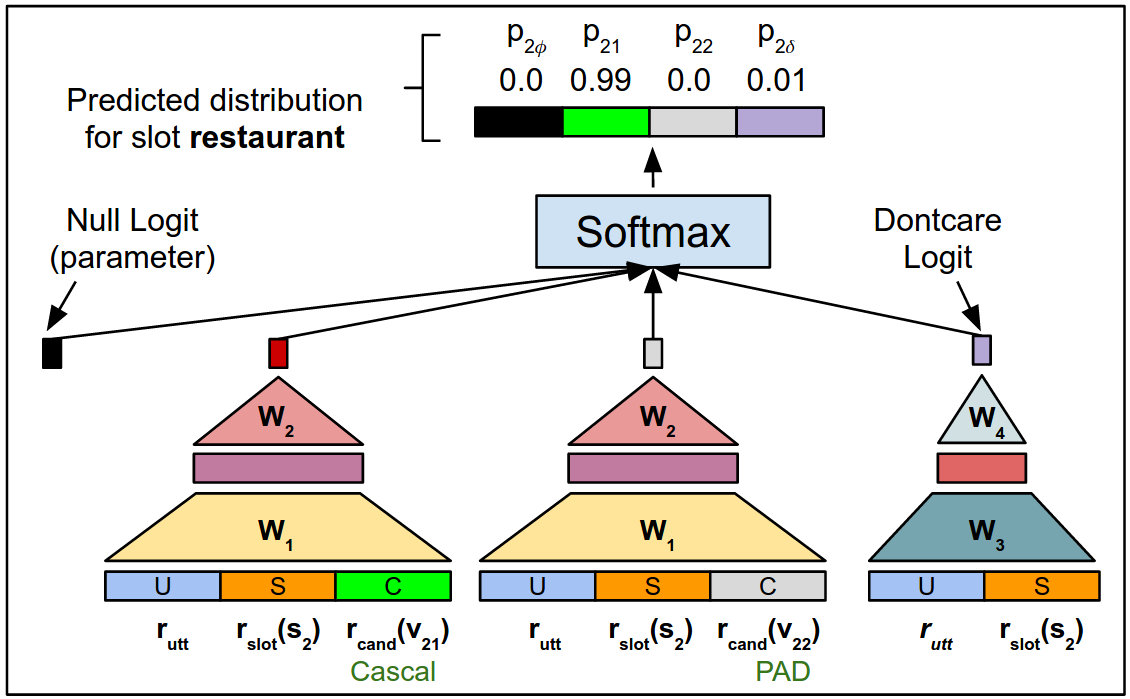}
  \end{subfigure}
  \caption{Candidate scorers for slots ``time" (shown in the top part
    of the figure) and ``restaurant" (shown in the bottom part of
    figure) after the final user turn in the example dialogue in
    Figure~\ref{fig:dialogue-example}. Note that, the figure is
    simplified to show the case where $K=2$. In cases the size of the
    candidate set is less than $K$, the remaining entries are padded
    with the {\em PAD} token. All neural network parameters are shared
    across these slots. The indicated candidate set values for each
    slot (e.g., ``6 pm'' and ``7 pm'' for ``time'' slot and
    ``Cascal'' for ``restaurant'' slot, shown below the network)
    are generated from system and user utterances in the dialogue. The
    figure is color-coded to show weights and features that are shared
    across different candidates and slots.}
  \label{fig:fig1}
  \vspace{-0.3cm}
\end{figure}

In order to keep the size of the distribution for a slot constant over
all turns, we add $K - |C_s^t|$ dummy values to ${V'}_s^t$ (i.e.,
padding, shown as {\em PAD} in Figure~\ref{fig:fig1}), giving a
fixed distribution of size $K + 2$. Furthermore, for most of the slots,
$|{V'}_s| = K + 2 \ll |V_s|$. This achieves a compact representation
of the distribution and is computationally efficient.

\section{Dialogue State Tracking}
This section describes the architecture of proposed neural network
models as well as extraction of domain, slot, and value independent
feature sets.

\subsection{Model Description}
The dialogue state tracker is a discriminative model which takes the candidate
set for each slot (such as restaurant, time, or \#people) as input and updates
the score for each candidate in the set. It also
identifies which slots have no constraints (e.g., \emph{dontcare}, where the
user specifically mentioned s/he is fine with any value for this slot) or
have not
been specified yet (\emph{null}). As mentioned in Section \ref{candset}, the
candidate
set is initialized using the current user utterance, the preceding system
utterance and the previous candidate set.  At user turn $t$, the DST model uses
the candidate set from the previous user turn (denoted as $C_s^{t-1}$) with
their scores, latest user and system utterances and their dialogue acts
to extract utterance
related ($r_{utt}^t$), slot related ($r_{slot}^t(s)$) and candidate related
($r_{cand}^t(c^t_{s, i})$) features. These features are then used by the
candidate scorers (for each slot) to update the score of each candidate in the
candidate set.
Let $g_s^t = r_{utt}^t \oplus
r_{slot}^t(s)$, $f_{c_{s, i}}^t = g_s^t \oplus r_{cand}^t(c^t_{s, i})$ and
$l_{\phi_s}^t = l_{\phi_s}$, where $\oplus$ denotes the concatenation of feature
vectors. The scores $p_\alpha^t$ for each $\alpha \in
{V'}_s^t = C^t_s \cup \{\delta_s, \phi_s\}$ are updated as shown in Figure
\ref{fig:fig1} using the equations:

\begin{equation} l_{c_{s, i}}^t = W^s_2 \cdot \sigma(W^s_1 \cdot f_{c_{s, i}}^t
  + b^s_1) + b^s_2
  \label{eq1}
\end{equation}
\begin{equation}
  l_{\delta_s}^t = W^s_4 \cdot \sigma(W^s_3 \cdot g_s^t + b^s_3) + b^s_4
  \label{eq2}
\end{equation}
\begin{equation}
  p_{\alpha}^t = \frac{\exp(l_\alpha^t)}{\exp(l_{\phi_s}^t) +
  \exp(l_{\delta_s}^t) + \Sigma_i\exp(l_{c_{s, i}}^t)}
  \label{eq3}
\end{equation}

Here $l_{\phi_s}$, $W_k^s$ and $b_k^s$ are trainable model parameters for $1
\leq k \leq 4$. Next, we describe how the features $r_{utt}^t$,  $r_{slot}^t(s)$
and $r_{cand}^t(c^t_{s, i})$ are computed.

\subsection{Feature Extraction}
\begin{figure*}[t]
  \centering
  \includegraphics[width=\linewidth]{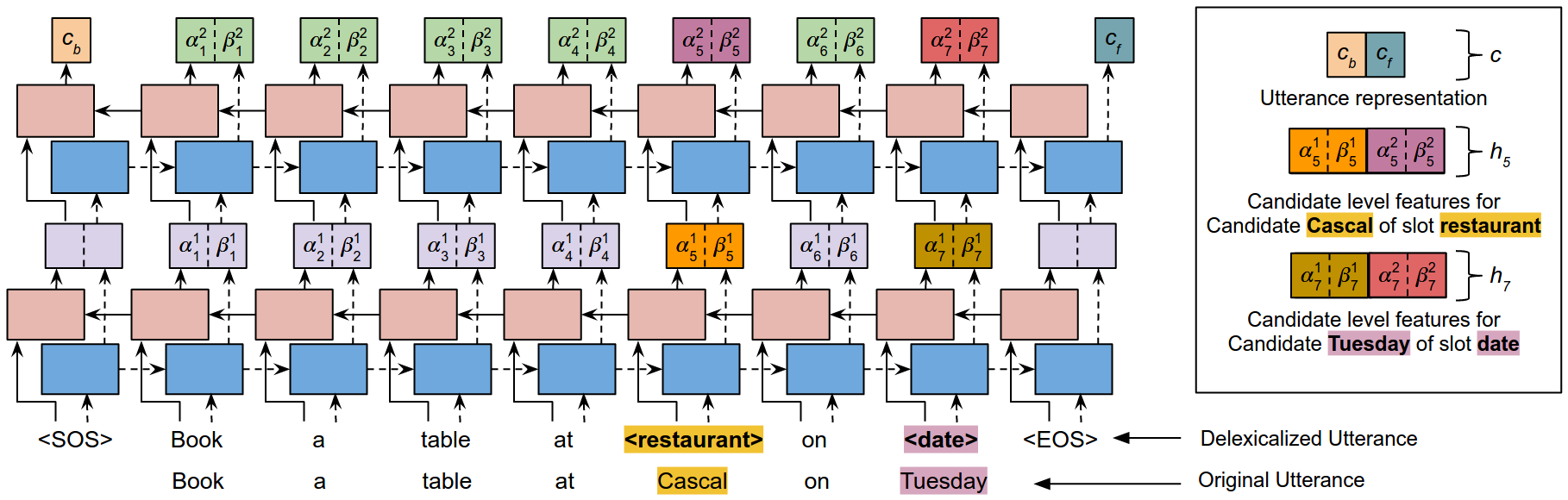}
  \caption{A two layer stacked bi-directional GRU network for feature
    extraction from system and user utterances. The final forward
    ($c_f$) and backward ($c_b$) states are used as the utterance
    representation ($c$). Candidate related features ($h_5$ and $h_7$
    for the two slot values in the delexicalized utterance: ``Cascal''
    for ``restaurant'' slot and ``Tuesday'' for ``date'' slot) are extracted
    from the two hidden layers of GRU cells located at the position of the
    candidate in the utterance. Their computation is depicted in the box on the
    right. Each cell is color-coded to match the ones in the GRU network for
    easy detection.}
  \label{fig:fig2}
\end{figure*}
Following earlier work on neural network based models for
DST~\cite{henderson2014, wen2016}, our model uses \emph{delexicalized}
utterances for extracting features. An utterance is
\emph{delexicalized} by substituting all the values associated with
the slot $s$, as recognized by LU, with a special token $delex(s)$.
For delexicalization, we use a deep learning based LU model\footnote{Not jointly
trained with DST.} (as described in~\cite{ankur:2017}) which does not need
knowledge of all values associated with a slot or rely on manually constructed
dictionaries. Furthermore, we do not delexicalize slot names from the utterances,
only the slot values are delexicalized.

The \emph{delexicalized} utterance at turn $t$ is fed to a two layer stacked
bi-directional GRU network (Figure \ref{fig:fig2}). The final forward state
($c_f^t$) and backward state ($c_b^t$) can be treated as a compact
representation of the whole utterance, whereas the states ($h_k^t$ for token at
index $k$ formed by concatenating the forward and backward states) encode the
context around a token\footnote{Note that the superscript $^t$, as in $c_f^t$,
is used to denote features for the examples at turn $t$}.
This approach is similar to \cite{wen2016}, which uses a convolutional
neural network (CNN) to obtain these features. The motivation for using
recurrent neural networks is to capture long term dependencies, instead of the
CNN features that aim to encode the local context of each token.

In addition to the tagging the values of slots, our LU also predicts the speech
acts corresponding to the user utterance, e.g., \emph{affirm},
\emph{negate(time)}, etc. These speech acts may have an optional slot parameter,
if a slot can be deduced from the utterance. For example, the utterance  ``No"
corresponds to a \emph{negate} act, whereas the utterance ``That time
doesn't work for me." corresponds to \emph{negate(time)}.

To capture information from the preceding system turn, we also extract features
from the system utterances and their dialogue acts as output by the dialogue
policy module. Some examples of system dialogue acts are \emph{goodbye},
\emph{request(time)} and \emph{inform(time=``6 pm")}. The \emph{delexicalized}
versions of system utterances are obtained from the language generation
component of the dialogue system.

\subsubsection{Utterance related features}
Utterance related features, $r_{utt}$, are relevant to all the
candidates for all slots in the task domain and hence are shared
across all candidate scorers. For example, in Figure \ref{fig:fig1},
$r_{utt}$ is the same for all candidate values of the ``restaurant'' and
``time'' slots.  The utterance related features are defined as

\begin{equation}
  r_{utt}^t =  c^t \oplus a_{u}^t \oplus {c'}^t \oplus {a'}_{u}^t
  \label{utt_feat}
\end{equation}
where $c^t$ is the user utterance representation obtained by concatenating the
final forward and backward states of the bi-directional GRU network (Figure
\ref{fig:fig2}) and $a_u^t$ is binary vector denoting the presence of
dialogue acts which don't have any slot or value argument (such as, ``greeting''
and ``negate'') in the user's utterance (as estimated by the LU).
${c'}^t$ and
${a'}_u^t$ denote the corresponding features for the system utterance preceding
the user's utterance.

\subsubsection{Slot related features}
The slot related features, $r_{slot}$, are relevant to a particular
slot and are shared across all candidate scorers for a
slot. For example, see Figure \ref{fig:fig1}, where $r_{slot}$ are the
same for ``time=6pm'', ``time=7pm'', ``time=dontcare''. The slot
related features for a slot $s$ are defined as
\begin{equation}
  r_{slot}^t(s) = a^t_{s}(s) \oplus {a'}^t_{s}(s) \oplus p_{\delta_s}^{t-1}
  \oplus p_{\phi_s}^{t-1}
  \label{slot_feat}
\end{equation}
where $p_{\delta_s}^{t-1}$ and $p_{\phi_s}^{t-1}$ are the estimated scores of
the special values \emph{dontcare} and \emph{null} respectively in the previous
turn's DST output and $a^t_{s}(s)$ is a binary vector denoting the presence of
dialogue acts having slot $s$ as the argument, e.g., \emph{request($s$)},
\emph{deny($s$)} in the user turn. ${a'}^t_{s}(s)$ is the corresponding binary
vector for the system turn
dialogue acts.

\subsubsection{Candidate related features}
The candidate related features, $r_{cand}$, are relevant to a
particular value in the candidate set for a slot and are not shared with any
other candidate
scorer. For a candidate $\hat c = c^t_{s, i}$ for the $i^{th}$ candidate in the
candidate set for a slot $s$, the
candidate related features are defined as
\begin{equation}
  r_{cand}^t(\hat c) = a_{c}^t(\hat c) \oplus {a'}_{c}^t(\hat c) \oplus p_{\hat
  c}^{t-1} \oplus \sum_{k \in T} h^t_{k} \oplus \sum_{k \in T'}{h'}^t_{k}
  \label{cand_feat}
\end{equation}
where $a_{c}^t(\hat c)$ is a binary vector denoting the presence of
all system dialogue acts associated with the candidate $\hat c$, e.g.,
the feature corresponding to act \emph{inform} has a value of 1,
whereas features corresponding to all other acts have a value of 0 for
the candidate with value \emph{``cheap"} for the slot \emph{price} due
to the system intent \emph{inform(price=``cheap")}. $p_{\hat c}^{t-1}$
is the predicted score of the candidate in the DST output for previous
user turn or $0$ if $\hat c$ was not present in $C^{t-1}_s$. $T$ is
the set of indices in the \emph{delexicalized} system utterance at
which the candidate was present prior to delexicalization. $h_k^t, k
\in T$, are the hidden state vectors obtained from utterance feature
extraction network (Figure \ref{fig:fig2}) and encode the context in
the utterance for the candidate $\hat c$. If $\hat c$ is not present
in the user utterance, $T$ is empty and the result of the summation is
taken to be a vector of zeros of the appropriate
size. ${a'}_{c}^t(\hat c)$, $T'$ and ${h'}^t_k$ are the corresponding
values for the system utterance.

\subsection{Parameter Sharing and Transfer Learning}

One of the main challenges for supervised learning based DST
approaches is the difficulty of creating labeled dialogue datasets.
Typically, in a multi-domain setting, one would train
separate models for each domain, thus requiring labeled training data for each
new domain. Sharing or transferring model parameters from one domain to another
reduces the requirement for labeled training data.

In the above formulation, the candidate scorer parameters $l_{\phi_s}$, $W_k^s$
and $b_k^s$, $1 \leq k \leq 4$, are defined per slot, whereas the GRU network
parameters are defined for the domain. The dimensions of these parameters don't
depend on the slot and domain respectively and thus allows us to do parameter
sharing or transfer across different domains.
In our experiments, we investigate sharing parameters across all slots within a
domain, transferring model parameters to a new domain without using any
in-domain data and joint training with shared parameters across different
domains.



\section{Evaluation}
\begin{table}[t]
  \caption{ Datasets used for training and evaluation. The number of dialogues
  is given for train, dev and test sets respectively. The last column
  presents the percentage of slot values in the test set that were not observed in the training data.}
  \label{tab:datasets}
  \centering
  \begin{tabular}{l l c c}
    \toprule
    \multicolumn{1}{c}{\textbf{Dataset}}& \multicolumn{1}{c}{\textbf{Slots}} &
    \multicolumn{1}{c}{\textbf{\# Dialogues}} & {\bf OOV \%}\\
    \midrule
     \rowcolor{noshadecolor}DSTC2 & \specialcell{pricerange, area,\\ food} &
    1612, 506, 1117 & 4.7\% \\
   \rowcolor{shadecolor}Sim-R &
    \specialcell{pricerange, area,\\ restaurant, food, \\ \#people,
      meal,\\ date, time}  &
    1116, 349, 775 & 43.0\% \\
  \rowcolor{noshadecolor}Sim-M & \specialcell{date, time, movie,
                                              \\theatre, area }&
    384, 120, 264 & 45.8\% \\
    \bottomrule
  \end{tabular}
  \vspace{-0.4cm}
\end{table}
The output of DST is a distribution of probabilities for candidate values of
each slot. To calculate the slot assignments, the value(s) with probability
above a threshold (tuned using dev set) are chosen. We use joint goal accuracy
as the metric for evaluation. This metric compares the predicted slot
assignments to the ground truth at every dialogue turn, and the output is
considered correct only if all the predicted slot values exactly match the
ground truth values.

\subsection{Datasets}
We obtain dialogues from two separate datasets spanning over the restaurant and
movie domains:
\begin{enumerate}
\itemsep0em
\item \textbf{DSTC2} (restaurant): We use the top ASR
  hypothesis, all LU hypotheses, system dialogue acts, the system
  utterance and turn level state labels for informable slots.
\item \textbf{Simulated Dialogues}(restaurant and movie): To
  experiment with slot scalability, we formed new goal oriented
  dialogue data sets restaurant (Sim-R) and movie (Sim-M)
  domains. These conversations are generated using an agenda-based
  user simulator~\cite{SchatzmannEtAl} interacting with a rule based
  dialogue policy. The generated utterances are then paraphrased by
  crowd-workers. We use the system dialogue acts, paraphrased user and
  system utterances, LU outputs for each user utterance, and turn
  level state labels. These datasets are available at
  \href{http://github.com/google-research-datasets/simulated-dialogue/}
       {\texttt{github.com/google-research-datasets/\\simulated-dialogue}}.
\end{enumerate}
Table~\ref{tab:datasets} summarizes the number of dialogues in the
training, development and test sets of each dataset. When splitting
the simulated datasets, to show the scalability of our approach, we
formed dev and test sets with conversations that include slot values that were
not observed in the training set. Table~\ref{tab:datasets}
also lists percentages of slot values in the test set that were
not observed in the training set (i.e., OOV
column). Note that the OOV slot rates in the simulated conversations
are significantly higher than the DSTC2 conversations.

\subsection{Experiments}
For training, we use mini-batch SGD with Adam optimizer~\cite{kingma2014adam}
to minimize the sum of cross-entropy loss for all the slots, backpropagating
through all the candidate scorers, the GRU feature extraction network and the
token embeddings. For each experiment, we used the dev set to identify the best
value (sampled within the specified range) for token embedding dimension(50-100), GRU
state size(50-100) and learning rate(0.001-0.1) using grid search.

We present two groups of experimental results. The first set of
experiments compare the impact of model architectures and weight
sharing on the three datasets. Table~\ref{tab:dstc2} presents joint
goal accuracy results with a rule-based system as described
in~\cite{DSTC2} as the baseline. The two results columns in this table
show results with no weight sharing between slots (i.e., weights are
trained per slot type) and with weight sharing across all slots in
each domain. For each dataset, results with CNN and GRU for
representing system and user utterances is also presented. For both
DSTC2 and simulated restaurant domain conversations, the GRU model
with weight sharing resulted in better performance. The results are
mixed with the movie domain conversations, but the movie dataset is
also much smaller than the other two. In the rest of the experiments,
we present results with weight sharing and GRUs.

 \begin{table}[t]
   \caption{ Joint goal accuracy results on different datasets to show the
             effect of slot level parameter sharing and comparison of
             Convolutional vs GRU based feature extraction.}
  \label{tab:dstc2}
  \centering
  \begin{tabular}{c l  c c}
    \toprule
      \textbf{Dataset} & \specialcell{\textbf{Feature} \\ \textbf{Extraction}} &
      \textbf{Not Shared} & \textbf{Shared} \\
      \midrule
      \multirow{3}{*}{DSTC2}
          & Baseline &   -   & 0.619 \\
          & CNN      & 0.697 & 0.695 \\
          & GRU      & 0.666 & {\bf 0.703} \\
          \rowcolor{shadecolor}
          & Baseline &   -   & 0.872 \\
          \rowcolor{shadecolor}
	  & CNN      & 0.892 & {\bf 0.944} \\
          \rowcolor{shadecolor}
          \multirow{-3}{*}{Sim-R}
          & GRU      & 0.878 & 0.937 \\
          \multirow{3}{*}{Sim-M}
          & Baseline &   -   & 0.967 \\
          & CNN      & {\bf 0.968} & 0.959 \\
          & GRU      & 0.959 & 0.945 \\
    \bottomrule
  \end{tabular}
  \vspace{-0.3cm}
\end{table}

The second set of experiments (Table~\ref{tab:multidomain}) evaluate sharing and
transfer of parameters to different datasets/domains, which is the main
contribution of our
work. In this setting parameters are always shared across all slots
during training and are replicated for each slot in the eval dataset
during evaluation. For this experiment, we use the transcript
utterances in DSTC2 training set because ASR noise is absent in the
other two datasets. We observe that the model parameters are
transferable to another dataset which has a different set of
slots. For Sim-M, we are able to achieve a good performance without
using any in-domain data. Adding in-domain data improves the
performance further and a combination of all datasets beats the best
performing model trained just on a single dataset on both movie and
restaurant datasets. Obtaining labeled training data for DST is a
tedious task and thus the ability to leverage existing datasets is
very important.

 \begin{table}[t]
   \caption{ Joint goal accuracy results for transfer of parameters to a new
             domain. Transcript utterances from DSTC2 training set are used
             (denoted by DSTC2*).}
  \label{tab:multidomain}
  \centering
  \begin{tabular}{l l c}
    \toprule
    \textbf{Train} & \textbf{Eval} & \textbf{Joint Goal Acc.} \\
    \midrule
    DSTC2*                 & Sim-R & 0.877 \\
    Sim-M                  & Sim-R & 0.855 \\
    DSTC2* + Sim-M         & Sim-R & 0.918 \\
    DSTC2* + Sim-M + Sim-R & Sim-R & \textbf{0.953} \\
    \rowcolor{shadecolor}
    DSTC2*                 & Sim-M & 0.940 \\
    \rowcolor{shadecolor}
    Sim-R                  & Sim-M & 0.971 \\
    \rowcolor{shadecolor}
    DSTC2* + Sim-R         & Sim-M & 0.972 \\
    \rowcolor{shadecolor}
    DSTC2* + Sim-R + Sim-M & Sim-M & \textbf{0.974} \\
    \bottomrule
  \end{tabular}
\end{table}

\subsection{Discussion}

Table~\ref{tab:dd} compares our approach with the previous work
on DSTC2. We present 3 results from the work of Mrksic et al.~\cite{mrksic2016},
showing their baseline on the second row and their improvements.
While the architecture in the proposed approach is actually comparable to the
baseline in~\cite{mrksic2016}, more elaborate DST architectures like
NBT-CNN~\cite{mrksic2016} can give better results on DSTC2.
However, the advantage of the proposed approach is ability to handle slots with
unbounded value sets and transfer learning to new slots and domains.
 \begin{table}[t]
   \caption{ Comparison with other approaches on DSTC2. }
  \label{tab:dd}
  \centering
  \begin{tabular}{l c}
    \toprule
    \textbf{DST Model} & \textbf{Joint Goal Acc.} \\
    \midrule
    Rule-based Baseline~\cite{DSTC2} & 0.619 \\
    Delexicalization-based Model~\cite{mrksic2016} & 0.691 \\
    \specialcell{Delexicalization-based Model \\ + Semantic Dictionary~\cite{mrksic2016}} & 0.729 \\
    Neural Belief Tracker (NBT-CNN)~\cite{mrksic2016} & 0.734 \\
    Scalable Multi-domain DST & 0.703 \\
    \bottomrule
  \end{tabular}
\end{table}

\section{Conclusions}

Three challenges with the state-of-the art models for DST are handling
unobserved slot values, dealing with very large or
unbounded value sets and building models in new domains due to the
cost of building new annotated data sets.  In this paper, we
introduced a novel approach for DST that derives a
candidate set of slot values from dialogue history and possibly
external knowledge, and restricts state tracking to this set for
efficiency. The proposed approach can handle slots with large or unbounded value
sets. In experimental results, we show that the new approach benefits from
sharing of model parameters across the slots of each
domain. Furthermore, model parameters can be transferred to a
new domain which allows for transfer learning and bootstrapping
of high performance models in new domains.

\noindent
{\bf Acknowledgments:} We would like to thank Matthew Henderson for useful
feedback and advice regarding presentation.


\bibliographystyle{IEEEbib}
\bibliography{strings,refs}

\begin{thebibliography}{10}

\bibitem{henderson2013deep}
Matthew Henderson, Blaise Thomson, and Steve Young,
\newblock ``Deep neural network approach for the dialog state tracking
  challenge,''
\newblock in {\em Proceedings of the SIGDIAL 2013 Conference}, 2013, pp.
  467--471.

\bibitem{shah2016interactive}
Pararth Shah, Dilek Hakkani-T{\"u}r, and Larry Heck,
\newblock ``Interactive reinforcement learning for task-oriented dialogue
  management,''
\newblock in {\em NIPS 2016 Deep Learning for Action and Interaction Workshop},
  2016.

\bibitem{DSTC1}
J.~Williams, A.~Raux, D.~Ramachandran, and A.~Black,
\newblock ``The dialog state tracking challenge,''
\newblock in {\em Proceedings of the SIGDIAL 2013 Conference}, 2013, pp.
  404--413.

\bibitem{DSTC2}
M.~Henderson, B.~Thomson, and J.~Williams,
\newblock ``The second dialog state tracking challenge,''
\newblock in {\em 15th Annual Meeting of the Special Interest Group on
  Discourse and Dialogue}, 2014, vol. 263.

\bibitem{DSTC3}
M.~Henderson, B.~Thomson, and J.~Williams,
\newblock ``The third dialog state tracking challenge,''
\newblock in {\em Spoken Language Technology Workshop (SLT), 2014 IEEE}. IEEE,
  2014, pp. 324--329.

\bibitem{DSTC4}
Seokhwan Kim, Luis~Fernando D'Haro, Rafael~E. Banchs, Jason Williams, and
  Matthew Henderson,
\newblock ``{The Fourth Dialog State Tracking Challenge},''
\newblock in {\em Proceedings of the 7th International Workshop on Spoken
  Dialogue Systems (IWSDS)}, 2016.

\bibitem{DSTC5}
Seokhwan Kim, Luis~Fernando D'Haro, Rafael~E. Banchs, Jason Williams, Matthew
  Henderson, and Koichiro Yoshino,
\newblock ``{The Fifth Dialog State Tracking Challenge},''
\newblock in {\em Proceedings of the 2016 IEEE Workshop on Spoken Language
  Technology (SLT)}, 2016.

\bibitem{henderson2014}
M.~Henderson, B.~Thomson, and S.~Young,
\newblock ``Word-based dialog state tracking with recurrent neural networks,''
\newblock in {\em Proceedings of the 15th Annual Meeting of the Special
  Interest Group on Discourse and Dialogue (SIGDIAL)}, 2014, pp. 292--299.

\bibitem{mrksic2015}
N.~Mrk{\v{s}}i{\'c}, D.~S{\'e}aghdha, B.~Thomson, M.~Ga{\v{s}}i{\'c}, P.-H. Su,
  D.~Vandyke, T.\-H. Wen, and S.~Young,
\newblock ``Multi-domain dialog state tracking using recurrent neural
  networks,''
\newblock {\em arXiv preprint arXiv:1506.07190}, 2015.

\bibitem{mrksic2016}
N.~Mrk{\v{s}}i{\'c}, D.~S{\'e}aghdha, T.\-H. Wen, B.~Thomson, and S.~Young,
\newblock ``Neural belief tracker: Data-driven dialogue state tracking,''
\newblock {\em arXiv preprint arXiv:1606.03777}, 2016.

\bibitem{wen2016}
Tsung-Hsien Wen, David Vandyke, Nikola Mrksic, Milica Gasic, Lina~M
  Rojas-Barahona, Pei-Hao Su, Stefan Ultes, and Steve Young,
\newblock ``A network-based end-to-end trainable task-oriented dialogue
  system,''
\newblock {\em arXiv preprint arXiv:1604.04562}, 2016.

\bibitem{hori2016dialog}
Takaaki Hori, Hai Wang, Chiori Hori, Shinji Watanabe, Bret Harsham, Jonathan
  Le~Roux, John~R Hershey, Yusuke Koji, Yi~Jing, Zhaocheng Zhu, et~al.,
\newblock ``Dialog state tracking with attention-based sequence-to-sequence
  learning,''
\newblock in {\em Spoken Language Technology Workshop (SLT), 2016 IEEE}. IEEE,
  2016, pp. 552--558.

\bibitem{dernoncourt2017robust}
Franck Dernoncourt, Ji~Young Lee, Trung~H Bui, and Hung~H Bui,
\newblock ``Robust dialog state tracking for large ontologies,''
\newblock in {\em Dialogues with Social Robots}, pp. 475--485. Springer, 2017.

\bibitem{BingLiu2017}
Bing Liu and Ian Lane,
\newblock ``An end-to-end trainable neural network model with belief tracking
  for task-oriented dialog,''
\newblock in {\em Proceedings of Interspeech}, 2017.

\bibitem{dilek:2016}
D.~Hakkani-T{\"u}r, G.~Tur, A.~Celikyilmaz, Y.-N. Chen, J.~Gao, L.~Deng, and
  Y.-Y. Wang,
\newblock ``Multi-domain joint semantic frame parsing using bi-directional
  rnn-lstm,''
\newblock in {\em Proceedings of Interspeech}, 2016.

\bibitem{jaech:2016}
A.~Jaech, L.~Heck, and M.Ostendorf,
\newblock ``Domain adaptation of recurrent neural networks for natural language
  understanding,''
\newblock in {\em Proceedings of Interspeech}, 2016.

\bibitem{ankur:2017}
A.~Bapna, D.~Hakkani-T{\"u}r, and L.~Heck,
\newblock ``Towards zero-shot frame semantic parsing for domain scaling,''
\newblock in {\em In Submission.}, 2017.

\bibitem{williams2013multi}
Jason Williams,
\newblock ``Multi-domain learning and generalization in dialog state
  tracking,''
\newblock in {\em Proceedings of SIGDIAL}. Citeseer, 2013, vol.~62.

\bibitem{shi2016}
Hongjie Shi, Takashi Ushio, Mitsuru Endo, Katsuyoshi Yamagami, and Noriaki
  Horii,
\newblock ``A multichannel convolutional neural network for cross-language
  dialog state tracking,''
\newblock in {\em Proceedings of IEEE Spoken Language Technology Workshop
  (SLT)}, 2016, pp. 559--564.

\bibitem{williams2014web}
Jason~D Williams,
\newblock ``Web-style ranking and slu combination for dialog state tracking,''
\newblock in {\em Proceedings of the 15th Annual Meeting of the Special
  Interest Group on Discourse and Dialogue (SIGDIAL)}, 2014, pp. 282--291.

\bibitem{williams2005scaling}
Jason~D Williams and Steve Young,
\newblock ``Scaling up pomdps for dialog management: The``summary
  pomdp''method,''
\newblock in {\em Automatic Speech Recognition and Understanding, 2005 IEEE
  Workshop on}. IEEE, 2005, pp. 177--182.

\bibitem{SchatzmannEtAl}
Jost Schatzmann, Blaise Thomson, Karl Weilhammer, Hui Ye, and Steve Young,
\newblock ``Agenda-based user simulation for bootstrapping a pomdp dialogue
  system,''
\newblock in {\em Human Language Technologies 2007: The Conference of the North
  American Chapter of the Association for Computational Linguistics (NAACL)},
  2007, pp. 149--152.

\bibitem{kingma2014adam}
Diederik~P. Kingma and Jimmy Ba,
\newblock ``Adam: A method for stochastic optimization,''
\newblock in {\em Proceedings of the 3rd International Conference on Learning
  Representations (ICLR)}, 2014.

\end{thebibliography}

\end{document}